\def\year{2022}\relax
\documentclass[letterpaper]{article} 
\usepackage{aaai22}  
\usepackage{times}  
\usepackage{helvet}  
\usepackage{courier}  
\usepackage[hyphens]{url}  
\usepackage{graphicx} 
\usepackage[caption=false, font=footnotesize]{subfig}
\usepackage{natbib}  
\usepackage{caption} 
\DeclareCaptionStyle{ruled}{labelfont=normalfont,labelsep=colon,strut=off} 
\usepackage{algorithm}

\usepackage{amsmath}
\usepackage{amsfonts} 
\usepackage{array}
\usepackage{booktabs}
\usepackage{algorithm}
\usepackage{algpseudocode}
\usepackage{color}
\usepackage{lscape}
\usepackage{url}
\usepackage{bbm}
\usepackage{soul}

\usepackage{amsmath}
\usepackage{amssymb}
\usepackage{graphicx}
\usepackage{amsmath,epsfig,amssymb,dsfont,amsthm}
\graphicspath{{figs/}}
\usepackage{caption}
\usepackage{subfig}
\usepackage{hhline}
\newcommand{\RR}{\ensuremath{\mathbb R}}
\newcommand{\minimize}[2]{\ensuremath{\underset{\substack{{#1}}}{\operatorname{minimize}}\quad#2}}

\newtheorem{lemma}{Lemma}

\DeclareMathAlphabet{\mathcal}{OMS}{cmsy}{m}{n}

\usepackage{url}

\usepackage[nomargin,inline,draft]{fixme}
\fxsetup{theme=color,mode=multiuser}
\FXRegisterAuthor{hpm}{ahpm}{\color{blue}hermina}
\FXRegisterAuthor{me}{ame}{\color{pink}mireille}
\FXRegisterAuthor{pf}{apf}{\color{red}pascal}
\FXRegisterAuthor{gc}{agc}{\color{orange}giovanni}

\usepackage{newfloat}
\usepackage{listings}
\floatstyle{ruled}
\newfloat{listing}{tb}{lst}{}
\floatname{listing}{Listing}
\title{fGOT: Graph Distances based on Filters and Optimal Transport}
\author {
    Hermina Petric Maretic\textsuperscript{\rm 1},
    Mireille El Gheche\textsuperscript{\rm 2},
    Giovanni Chierchia\textsuperscript{\rm 3},
    Pascal Frossard\textsuperscript{\rm 1}
}
\affiliations {
    \textsuperscript{\rm 1} EPFL, LTS4, Lausanne, Switzerland\\
  \textsuperscript{\rm 2} Sony AI, Z\"urich, Switzerland\\
    \textsuperscript{\rm 3} Université Gustave Eiffel, LIGM (UMR 8049),
  ESIEE Paris, CNRS, F-93162, Noisy-le-Grand, France\\
    hermina.petricmaretic@epfl.ch, mireille.elgheche@sony.com, giovanni.chierchia@esiee.fr, pascal.frossard@epfl.ch
}

\usepackage{bibentry}

\begin{document}
\maketitle
\begin{abstract}

Graph comparison deals with identifying similarities and dissimilarities between graphs. A major obstacle is the unknown alignment of graphs, as well as the lack of accurate and inexpensive comparison metrics. In this work we introduce the \textit{filter graph distance}. It is an optimal transport based distance which drives graph comparison through the probability distribution of filtered graph signals. 
This creates a highly flexible distance, capable of prioritising different spectral information in observed graphs, offering a wide range of choices for a comparison metric. We tackle the problem of graph alignment by computing graph permutations that minimise our new filter distances, which implicitly solves the graph comparison problem. 
We then propose a new approximate cost function that circumvents many computational difficulties inherent to graph comparison and permits the exploitation of fast algorithms such as mirror gradient descent, without grossly sacrificing the performance. We finally propose a novel algorithm derived from a stochastic version of mirror gradient descent, which accommodates the non-convexity of the alignment problem, offering a good trade-off between performance accuracy and speed. The experiments on graph alignment and classification show that the flexibility gained through filter graph distances can have a significant impact on performance, while the difference in speed offered by the approximation cost makes the framework applicable in practical settings. 

\end{abstract}

\section{Introduction}
\label{sec:fgot_intro}
With the rapid development of digitization in various domains, the volume of data continuously increases. A large share of it takes the form of structured data, which is often represented by graphs that capture potentially complex information structures. The analysis of such data often requires the design of algorithms that are able to properly compute distances between different graphs. It stays however pretty challenging to compare graphs for two main reasons. First, it is usually not clear \textit{a priori} how to align the nodes of graphs under comparison, leading to a very large number of possible solutions. Second, even if the nodes are aligned, it is not obvious how to define a meaningful metric for comparing graphs. In particular, a naive comparison of graph adjacency matrices is not an optimal measure, as it does not capture the importance of an edge in the graph. Edges can have a very different influence on the graph structure, as well as on its local or global spectral characteristics. 

In this paper, we propose the filter graph distance, a dissimilarity measure based on optimal transport, which compares two graphs through the probability distribution of data generated with graph filters. 
Graph filters have been largely studied in the field of Graph Signal Processing \cite{shuman_emerging_2013,ortega2018graph}, and offer a high level of flexibility in modelling the relationship between data and the underlying graph.\footnote{The data living on a graph is also referred to as a \emph{graph signal}.} Hence, their use in defining filter graph distance permits to capture a wide range of structural graph properties, including local characteristics, global structure, and any combination of spectral graph properties. 

%



Equipped with this new distance, we derive an efficient approximation for the graph alignment problem, which is a necessary step in graph comparison. Our approximation permits to remove computational complexity bottlenecks in the alignment problem (e.g., \cite{NIPS2019_9539}). It also enables the comparison of graphs of different size, which brings an important advantage in practice. The solution of the graph alignment problem directly leads to the computation of the minimal filter graph distance, which eventually permits to compare different graphs. 


Our new formulation of the graph alignment problem leads to a nonconvex optimization framework. We solve the problem efficiently with a novel stochastic algorithm based on mirror gradient descent (MDG), which has the peculiarity of being applicable to any filter graph distance. We demonstrate the benefits of our method in alignment, clustering and classification tasks with both synthetic and real graph datasets. Filter graph distances show better performance than the standard optimal transport distances on simple graph alignment tasks, while significantly decreasing the computational cost of alignment recovery. The proposed stochastic algorithm also achieves significantly better results for community detection in structured graphs, when compared to the vanilla MGD. This suggests that our algorithm successfully addresses the inherent difficulty of graph alignment. Finally, experiments on benchmark graph classification datasets demonstrate how the benefits of our method propagate to other distance-related tasks, confirming the importance of the flexibility introduced in our method.

The rest of this paper is structured as follows. The next section reviews the related work. Then we introduce the filter graph distances using the optimal transport framework and give a scalable approximation cost to the newly formulated optimal transport problem. After that, we propose a new stochastic algorithm for solving our new graph alignment problem via MGD. Finally, we asses the performance of the proposed approach in different tasks, where the benefits of the filter graph distance and the efficient stochastic algorithm is shown for synthetic and real graph datasets.



\section{Related work}
\label{sec:related_work}

Several works in the literature have studied the graph comparison problem. For example, graph alignment has been formulated as a quadratic assignment problem \cite{YanICMR2016,NIPS2017_6911}, both as a graph edit distance \cite{bougleux2017graph} or under the constraint that the solution is a permutation matrix. This results in an NP-hard problem, and several relaxations have been proposed to find approximate solutions to this problem \cite{cho2010reweighted,Zhou2012, NIPS2018_7365}. Recently, some works have studied the graph alignment problem from the perspective of optimal transport, which has inspired several advances in machine learning \cite{arjovsky2017wasserstein, peyre2019computational}. For example, 
\cite{GU201556} define a spectral distance by assigning a probability measure to the nodes via the spectrum representation of each graph and using Wasserstein distances between probability measures. This approach, however, disregards graph eigenvectors and thus does not take into account the full graph structure in the alignment problem. 
The authors in \cite{nikolentzos2017matching} propose to match the graph embeddings, where the latter are represented as bags of vectors, and the Wasserstein distance is computed between them. The authors also propose a heuristic to take into account possible node labels or signals. A Gumbel-sinkhorn network inspired by optimal transport was further proposed to infer permutations from data \cite{Mena2018, Emami2018}. The approach consists of producing a discrete permutation from a continuous doubly-stochastic matrix obtained with the Sinkhorn operator \cite{Sinkhorn1964}. More recently, optimal transport between stationary Markov chains has been used to find a coupling between graphs \cite{oconnor_graph_2021}.

Building on similar ideas, the work in \cite{memoli2011gromov} investigates the Gromov-Wasserstein distance for object matching, and the authors in \cite{2016-peyre-icml} propose an efficient algorithm to compute the Gromov-Wasserstein distance and the barycenter of pairwise dissimilarity matrices. The algorithm uses entropic regularization and Sinkhorn projections, as proposed by \cite{Cuturi2013}. Later, the authors in \cite{vayer2018optimal} build on this work to propose a distance for graphs and signals living on these graphs. The problem is given as a combination between the Gromov-Wasserstein of graph distance matrices and the Wasserstein distance of graph signals. Then, the work in \cite{xu2019gromov} proposes a method based on Gromov-Wasserstein which simultaneously learns the graph alignment and the embeddings of graph nodes. The node embeddings are derived using optimal transport, which in turn helps in the graph matching task. The authors follow up on \cite{NIPS2019_8569} by devising a scalable version of Gromov-Wasserstein distance for graph partitioning and matching. However, while the above methods solve the alignment problem using optimal transport, the simple distances between aligned graphs do not take into account the global structure of the graph, which is important for proper graph comparison. 

The graph optimal transport distance introduced by \cite{NIPS2019_9539, Petric2021Representing} has shown to successfully capture the global structure of graphs, representing their topology through smooth signal distributions. The work has been extended by \cite{maretic2020wasserstein} to accommodate for comparison of graphs of different sizes using a one-to-many framework, and by \cite{dong2020copt} for graph sketching. However, 
while smooth graph signals efficiently capture global graph properties, they can lack descriptiveness for other potentially interesting properties. 
Several graph distances based on heat diffusion have been proposed. \cite{hammond2013graph} propose a direct comparison of graph heat diffusion matrices, but stay limited to graphs of the same size. A spectral method proposed by \cite{tsitsulin2018netlsd} circumvents this problem by comparing heat kernel traces, but it still does not address the graph alignment problem. Closer to our work, a fast heat kernel distance based on optimal transport has recently been proposed by \cite{barbe2020graph}. The authors apply the Wasserstein distance between signals only after a graph filtering step, encoding the structural information of the graph into the filtered signals. However, this distance compares graphs through available signals, whereas our method uses the representation of graphs through signal distributions and therefore does not need to rely on actual signal availability. 
Furthermore, the above methods are limited only to heat diffusion models, while more general distances could often be of interest in practice.

\section{Filter Graph Alignment with Optimal Transport}
\label{sec:fgot_problem}

Despite recent advances in the analysis of graph data, it stays pretty challenging to define a meaningful distance between graphs. Instead of comparing graphs directly, we therefore propose to look at the signal distributions governed by graph filters. Specifically, we define the filter graph distance (fGOT) as a generalisation of the graph optimal transport (GOT) distance proposed by \cite{NIPS2019_9539}, which has the ability to emphasise specific spectral properties of the graph, such as high or low frequencies, local or global graph phenomena. We model these properties through filtered graph signals, which exploit the specific graph information and evenutally compare graphs through filtered signal distributions. 

\subsection{Preliminaries}

 Let $\mathcal{G} = (\mathcal{V}, \mathcal{E}, W)$ be an undirected, weighted graph with no labels and with a set of $N$ vertices $\mathcal{V}$, edges $\mathcal{E}$ and a weighted adjacency matrix $W$. The combinatorial graph Laplacian is defined as $L = D - W$, where $D$ is a diagonal matrix of node degrees. We define a signal on a graph as a vector $x \in \mathbb{R}^N$, where $x_n$ denotes the value of a signal $x$ on a vertex $n$.

We further define filtering in graph signal processing \cite{ortega2018graph} as $x_f = g(L) x$, where the filter $g(L)$ is an operator defined through the graph Laplacian matrix, and $x_f$ represents a filtered graph signal.\footnote{The choice of the filter $g(\cdot)$ drives different spectral characteristics of the graph signals. For example, a heat kernel $g(L) = e^{-\tau L}$ models a graph based on the nature of the spread of heat through the graph, usually emphasizing global graph properties. On the other hand, a high pass filter like $g(L)= L^2$ takes more local phenomena into account, prioritising high graph frequencies.} A graph filter can be represented in matrix form as
\begin{align}
g(L) &= U \hat{G} U^T \label{eq:filter_matrix}
\end{align}
with a diagonal matrix $\hat{G} = {\rm diag}\big(\hat{g}(\lambda_1),\dots,\hat{g}(\lambda_N)  \big)$,
where $U$ denotes the eigenvectors and $\lambda_i$ the eigenvalues of the Laplacian matrix $L$. 
Given a graph filter $g(L)$ and a Gaussian white noise signal $w \sim \mathcal{N}(0, I)$, we study random filtered graph signals given by $x_f =  g(L) w$.
They follow the Gaussian distribution
\begin{align}
\nu^{\mathcal{G}, g} := \mathcal{N}(0, g(L) I g(L)^T) = \mathcal{N}\big(0, g^2(L)\big), 
\end{align}
where $g(L)g(L)^T = g^2(L)$ follows from the fact that every graph filter $g(L)$ is symmetric by definition in Equation \ref{eq:filter_matrix}.

Our graph distance compares graphs through their respective filtered signal distributions. The choice of filter is critical to capture fine structural properties of graphs and to drive the resulting distance. For example, a distance based on the graph heat-kernel will model differences in the spread of heat through the graphs, emphasizing global dissimilarities.




\subsection{Filter graph distance}

Given two aligned graphs $\mathcal{G}_1$ and $\mathcal{G}_2$ of the same size\footnote{This assumption will be removed later.} with Laplacian matrices $L_1$ and $L_2$, we define the filter graph distance (fGOT) as the Wasserstein distance between the probabilistic distributions of their respective filtered signals, namely $\nu^{\mathcal{G}_1, g}$ and $\nu^{\mathcal{G}_2, g}$, defined earlier. In particular, the 2-Wasserstein distance measures the cumulative cost of transporting one distribution to the other in the optimal way and with respect to the Euclidean norm \cite{Monge_1781}.
For normal distributions such as $\nu^{\mathcal{G}_1, g}$ and $\nu^{\mathcal{G}_2, g}$, the 2-Wasserstein distance is equal to
\begin{align}\label{eq:wass_filter}
\mathcal{W}_2^2\big(\nu^{\mathcal{G}_1, g}, \nu^{\mathcal{G}_2, g}\big) &= 
{\rm Tr}\left(g^2(L_1)\right) + {\rm Tr}\left(g^2(L_2)\right) \nonumber \\ 
&-  2 \, {\rm Tr}\left(\sqrt{g(L_1)\, g^2(L_2) \, g(L_1)}\right).
\end{align}

\begin{figure}
	\centering
	\includegraphics[width=0.45\textwidth]{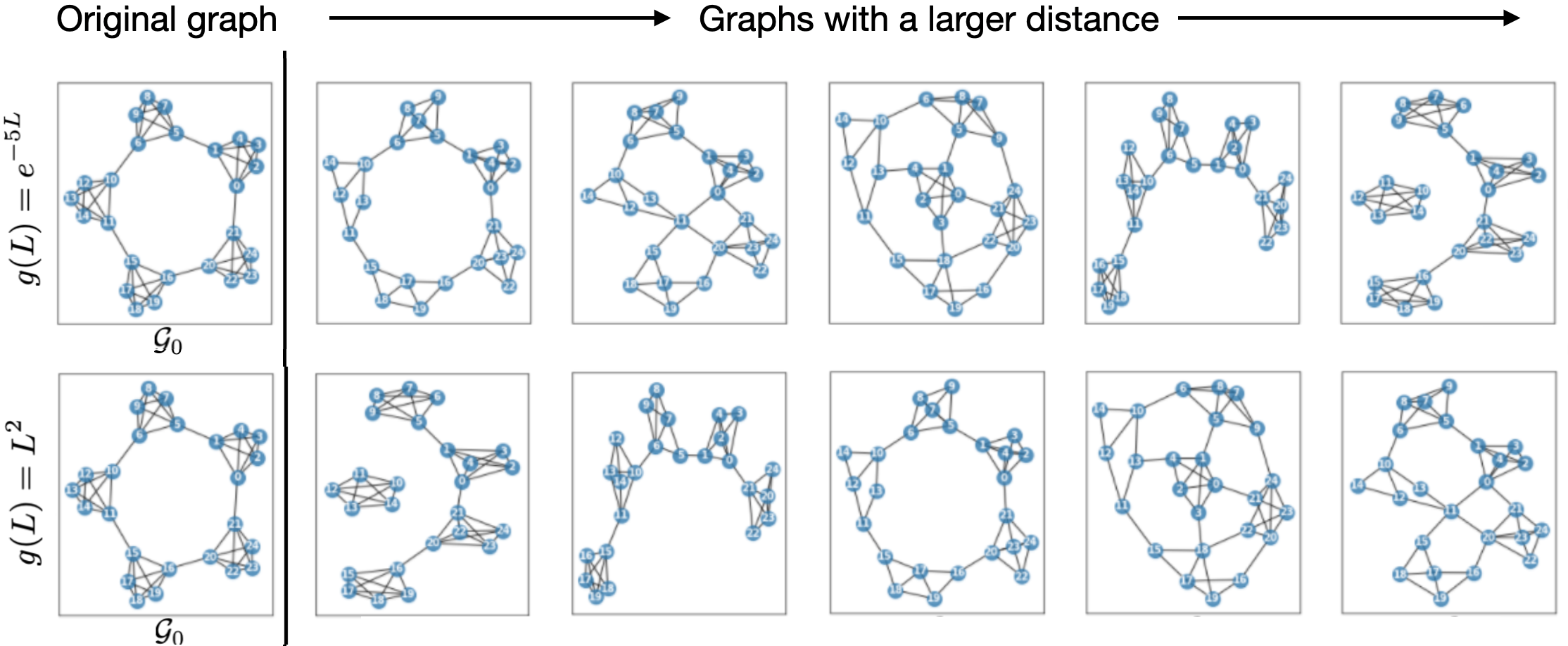}
	\caption{Graphs sorted by distance to $\mathcal{G}_0$ with respect to two different filter graph distances. The low-pass filter focuses on global graph properties (e.g., preservation of the ring structure, connectedness), while the high-pass filter distance considers local differences (e.g., immediate neighbourhood of each node, number of removed edges). Both can be of interest, depending on the application. For more details, see the Appendix.}
	\label{fig:fgot_example_simple}
\end{figure}

The fGOT distance compares the nature of graph filter responses, putting an emphasis on the specific properties of filtered signals. A simple example of this is shown in Figure \ref{fig:fgot_example_simple}. In particular, the smooth graph optimal transport (GOT) distance can be seen as a special case\footnote{Here $\dagger$ denotes a pseudoinverse operator.} of fGOT, with the low pass graph filter equal to $g(L) = \sqrt{L^\dagger}$.

\subsection{Alignment problem}

It is important to note that the distance introduced in \eqref{eq:wass_filter} requires knowledge of a correspondence between vertices of the two graphs being compared. While in some cases a consistent enumeration can be trivially chosen for all graphs, this requirement is often not realistic. Therefore, a proper alignment between graph vertices must be recovered for making graphs comparable. The alignment between two graphs $\mathcal{G}_1$ and $\mathcal{G}_2$ can be represented by a permutation matrix $P\in \RR^{N\times N}$ that maps a vertex of $\mathcal{G}_1$ to a vertex of $\mathcal{G}_2$. To take different alignments into account, we define a probability distribution for the permuted version of $\mathcal{G}_2$ as 
\begin{equation}
\nu_P^{\mathcal{G}_2, g} = \mathcal{N}\big(0, g^2(P L_2 P^T)\big).    
\end{equation} 
We then aim at searching for the optimal alignment that minimises the filter graph distance between $\mathcal{G}_1$ and a permuted version of $\mathcal{G}_2$, yielding the optimization problem defined as
\begin{equation}\label{eq:alig_filter}
	\minimize{P \in \mathcal{C}_{\rm perm}}{} \mathcal{W}_2^2\big(\nu^{\mathcal{G}_1, g}, \nu_P^{\mathcal{G}_2, g}\big) 
\end{equation}
where $\mathcal{C}_{\rm perm}$ denotes the set of permutation matrices
\begin{equation}\label{eq:perm_matrix}
	\mathcal{C}_{\rm perm} 
	=
	\left\{
	P \in \RR^{N\times N}:
	\;
	\begin{aligned}
	&(\forall i, \forall j)\; P_{ij} \in \{0,1\}
	\\ 
	&(\forall i)\; \textstyle \sum_{j} P_{ij} = 1
	\\
	&(\forall j)\; \textstyle \sum_{i} P_{ij} = 1
	\end{aligned}
	\right\}
	\!.
	\end{equation}

Solving Problem \eqref{eq:alig_filter} is computationally very challenging for several reasons, one of them being that the permutation $P$ appears under a matrix square root. For this reason, we propose to replace the criterion in Problem \eqref{eq:alig_filter} with an approximation through the upper bound $\widetilde{\mathcal{W}}_2^2$ of the filter graph distance, defined as
\begin{align}
\widetilde{\mathcal{W}}_2^2\big(\nu^{\mathcal{G}_1, g}, \nu^{\mathcal{G}_2, g}_{P}\big) 
&= {\rm Tr}\left(g^2(L_1)\right) + {\rm Tr}\left( g^2(L_2) \right) + \nonumber \\ 
&- 2\left\langle g(L_1)P g(L_2),  P \right\rangle.
\label{eq:fgot_surrogate}
\end{align} 
The theoretical justification is provided by the next lemmas. 



\begin{lemma}
\label{lemma_1}
	Let $P \in \mathcal{C}_{\rm perm}$ be a permutation matrix, and let $L \in \RR^{N \times N}$ be a graph Laplacian matrix. Then $g(PLP^T) = Pg(L)P^T$ for any graph filter $g(\cdot)$ defined as in (\ref{eq:filter_matrix}).
\end{lemma}
\begin{proof}
See the Appendix.
\end{proof}


\begin{lemma}
\label{lemma_2}
Let $\mathcal{G}_1$ and $\mathcal{G}_2$ be two graphs with their respective Laplacian matrices $L_1 \in \RR^{N \times N}$ and $L_2 \in \RR^{N \times N}$. Let the fGOT distance $\mathcal{W}_2^2$ be defined as in Equation \ref{eq:wass_filter}, and its approximation $\widetilde{\mathcal{W}}_2^2$ as in Equation \ref{eq:fgot_surrogate}. Then, for any graph filter $g(\cdot)$ defined as in (\ref{eq:filter_matrix}), and $P\in \mathcal{C}_{\rm perm}$ :
	\begin{align}
		\mathcal{W}_2^2\big(\nu^{\mathcal{G}_1, g}, \nu^{\mathcal{G}_2, g}_{P^T}\big) \leq \widetilde{\mathcal{W}}_2^2\big(\nu^{\mathcal{G}_1, g}, \nu^{\mathcal{G}_2, g}_{P}\big).
	\end{align}
\end{lemma}
\begin{proof}
See the Appendix.
\end{proof}
\begin{lemma}
\label{lemma_3}
Let $\mathcal{G}_1$ and $\mathcal{G}_2$ be two isomorphic graphs with their respective Laplacian matrices $L_1 \in \RR^{N \times N}$ and $L_2 \in \RR^{N \times N}$. Let the fGOT distance $\mathcal{W}_2^2$ be defined as in Equation \ref{eq:wass_filter}, and its approximation $\widetilde{\mathcal{W}}_2^2$ as in Equation \ref{eq:fgot_surrogate}. 
Then, for any graph filter $g(\cdot)$ defined as in (\ref{eq:filter_matrix}):
\begin{align}
	\min_{P\in \mathcal{C}_{\rm perm} } \mathcal{W}_2^2\big(\nu^{\mathcal{G}_1, g}, \nu^{\mathcal{G}_2, g}_{P^T}\big) = \min_{P\in \mathcal{C}_{\rm perm}}  \widetilde{\mathcal{W}}_2^2\big(\nu^{\mathcal{G}_1, g}, \nu^{\mathcal{G}_2, g}_{P}\big) = 0.
\end{align}
\end{lemma}
\begin{proof}
See the Appendix.
\end{proof}

In the next section, we propose an efficient algorithm to find an assignment matrix $P$ that minimizes the surrogate cost function in \eqref{eq:fgot_surrogate}, whose purpose is to approximately solve Problem \eqref{eq:alig_filter}. In addition to being computationally efficient, the proposed approach can advantageously handle the comparison of graphs with different sizes, because the cost function in Problem \eqref{eq:alig_filter} and its surrogate version in Equation \eqref{eq:fgot_surrogate} still hold when the assignment matrix $P$ is rectangular. 


\section{FGOT algorithm}
\label{sec:fgot_algo}

\subsection{Relaxed alignment problem}






Equipped with the surrogate fGOT distance in \eqref{eq:fgot_surrogate} to compare unaligned graphs, we now introduce the proposed formulation for approximate graph alignment. The last roadblock in solving the original graph alignment problem of Eq. (\ref{eq:alig_filter}) is the requirement that the sought solution must be a binary assignment matrix, since this criterion leads to a discrete optimisation problem with a factorial number of feasible solutions. We propose to circumvent this issue by relaxing the constraint through an implicit reformulation. Given two graphs $\mathcal{G}_1$ and $\mathcal{G}_2$ of possibly \emph{different} size, we represent a soft assignment between the graph nodes through a matrix of the following convex set
\begin{equation}\label{eq:fuzzy_assignement_constraint}
\mathcal{C}_{\rm fuzzy}
=
\left\{
P \in \RR^{|V_1| \times |V_2|}:
\;
\begin{aligned}
&(\forall i, \forall j)\; P_{ij} \geq 0
\\ 
&(\forall i)\; \textstyle \sum_{j} P_{ij}  = \frac{1}{|V_1|}
\\
&(\forall j)\; \textstyle \sum_{i} P_{ij} = \frac{1}{|V_2|}
\end{aligned}
\right\}
\!.
\end{equation}


Note that $P \in \mathcal{C}_{\rm perm}$ implies $P \in \mathcal{C}_{\rm fuzzy}$ up to a rescaling factor of $\sqrt{|V_1||V_2|}$. Using the surrogate fGOT distance in \eqref{eq:fgot_surrogate}, we can now express the approximate distance $\mathcal{D}$ between graphs $\mathcal{G}_1$ and $\mathcal{G}_2$ as the solution to the following graph alignment problem 
\begin{align}\label{eq:fgot_problem_with_fuzzy}
\mathcal{D}(\mathcal{G}_1, \mathcal{G}_2) = & \min_{P \in \mathcal{C}_{\rm fuzzy} }{\widetilde{\mathcal{W}}_2^2\big(\nu^{\mathcal{G}_1, g} , \nu^{\mathcal{G}_2, g}_{P} \big)} 
\end{align}

The standard approach to solve the above problem is by using the projected mirror gradient descent (MGD), where the projection onto $ \mathcal{C}_{\rm fuzzy}$ is computed according to the Kullback-Leibler ($\rm KL$) metric \cite{2015_IterativeBregman}. Adding the entropic regularisation $\epsilon H(P)$ to the problem yields the following iterative algorithm 
\begin{align}
P_{t+1} = \mathcal{P}_{\mathcal{C}_{\rm fuzzy}}^{\rm KL}\big(P_{t} \, \odot \,  e^{- \alpha q_t}\big),
\end{align}
where $q_t = \nabla \widetilde{\mathcal{W}}_2^2\big(\nu^{\mathcal{G}_1, g} , \nu^{\mathcal{G}_2, g}_{P_t} \big) - \epsilon \nabla H(P_t)$, the symbol $\odot$ is the Haddamard (pointwise) product between matrices, $\nabla$ represents the gradient operator, $\alpha$ is the step size and $\epsilon$ is the entropic regularisation parameter.
The KL projection $\mathcal{P}_{\mathcal{C}_{\rm fuzzy}}^{\rm KL}$ can be computed through the Sinkhorn operator $\mathcal{S}_\tau$ as \cite{Sinkhorn1964}, \cite{Cuturi2013}
\begin{equation}
\mathcal{P}_{\mathcal{C}_{\rm fuzzy}}^{\rm KL}(P) = \mathcal{S}_\tau (-\tau \log{P}),
\end{equation}
where $\tau>0$ is a small constant.

This is the standard approach commonly found in the literature. However, Problem \eqref{eq:fgot_problem_with_fuzzy} is non-convex and thus very susceptible to converge towards locally optimal solutions. To address this issue, we propose a novel stochastic version of MGD. 


\subsection{Proposed algorithm}

We reformulate Problem \eqref{eq:fgot_problem_with_fuzzy} with an implicit constraint, taking into account the KL projection appearing in the non-stochastic version of MGD algorithm. For the sake of clarity, we denote the KL projection to $\mathcal{C}_{\rm fuzzy}$ with 
$
\mathcal{B}(P) = \mathcal{P}_{\mathcal{C}_{\rm fuzzy}}^{\rm KL}(P),
$
and the optimization problem becomes
\begin{equation}\label{eq:fgot_problem_with_implicit}
\minimize{P \in \RR^{|V_1| \times |V_2|} }{\widetilde{\mathcal{W}}_2^2\big(\nu^{\mathcal{G}_1, g} , \nu^{\mathcal{G}_2, g}_{\mathcal{B}(P)} \big)}.
\end{equation}


\begin{algorithm}[t]
	\caption{Approximate solution to 
	Problem \eqref{eq:fgot_problem_with_fuzzy}}
	\label{alg:alg_mgd} 
	\begin{algorithmic}[1]
		\State{\bf Input:} {Graphs $\mathcal{G}_1$ and $\mathcal{G}_2$} 
		\State{\bf Input:} {Sampling $S\in\mathbb{N}$, step size $\alpha>0$ and $\tau>0$}
		\State{\bf Input:} {Setting matrix $\eta_0$ (constant) and $\sigma_0$ (random)}
		
		\For{$t=0,1,\dots$}
		\State Draw the samples $\epsilon_t^{(1)},\dots,\epsilon_t^{(N)}$ from $\mathcal{N}\big(0,I\big)$.
		
		\State Estimate the gradient $({\sf g}_t^{(\eta)}, {\sf g}_t^{(\sigma)})$ :
		$$
		\begin{aligned}
		{\sf g}_{t}^{(\eta)} &\approx \frac{1}{N}\sum_{n=1}^N \nabla \widetilde{\mathcal{W}}_2^2\big(\nu^{\mathcal{G}_1, g} , \nu^{\mathcal{G}_2, g}_{\mathcal{B}(P_n)} \big)\Big|_{P_n=\eta_t + \sigma_t\circ\epsilon_t^{(n)}} \\
		{\sf g}_{t}^{(\sigma)} &\!\!\approx \frac{1}{N}\sum_{n=1}^N \Big(\epsilon_t^{(n)} \!\circ\! \nabla \widetilde{\mathcal{W}}_2^2\big(\nu^{\mathcal{G}_1, g} , \nu^{\mathcal{G}_2, g}_{\mathcal{B}(P_n)} \big)\Big|_{P_n=\eta_t + \sigma_t\circ\epsilon_t^{(n)}}\Big)
		\end{aligned}
		$$
		\State Update $\eta_t$ and $\sigma_t$ using $({\sf g}_t^{(\eta)}, {\sf g}_t^{(\sigma)})$: 
		$$\begin{gathered}
		\eta_{t+1} = \eta_{t} - \alpha_t \sigma_{t}^2 \circ {\sf g}_{t}^{(\eta)}, 
		\quad
		\sigma_{t+1} = \sqrt{\sigma_{t}^2 + {\sf d}_{t}^2} - {\sf d}_{t}
		\end{gathered}$$
		\quad\; with ${\sf d}_{t} = \frac{1}{2}\alpha_t \sigma_{t}^2 \circ {\sf g}_{t}^{(\sigma)}$.
		\EndFor
		\State \textbf{Output:} $P = \mathcal{B}(\eta_*)$
	\end{algorithmic}
\end{algorithm}

Note that the KL projection is computed as a sequence of softmax operations, so the above cost function is constraint-free and differentiable \cite{Luise2018}, albeit nonconvex. To deal with nonconvexity, we propose to optimise the expectation of  $\widetilde{\mathcal{W}}_2^2\big(\nu^{\mathcal{G}_1, g} , \nu^{\mathcal{G}_2, g}_{\mathcal{B}(P)} \big)$ w.r.t. the parameters $\theta$ of some distribution $p_\theta$, that is 

\begin{equation}\label{eq:final_problem}
\operatorname*{minimize}_{\theta} \quad \underbrace{\mathbb{E}_{P\sim p_\theta}\big\{ \widetilde{\mathcal{W}}_2^2\big(\nu^{\mathcal{G}_1, g} , \nu^{\mathcal{G}_2, g}_{\mathcal{B}(P)} \big) \big\}}_{U(\theta)}.
\end{equation} 
The reformulated problem aims at shaping the distribution $p_\theta$ so as to put all its mass on a minimizer $P^*$ of $\widetilde{\mathcal{W}}_2^2$, thus integrating the use of Bayesian exploration in the optimization process. This approach is commonly used in many areas of stochastic search, such as evolution strategies \cite{Hansen2001, Wierstra2008, Salimans2017}.


Since the cost function in Problem \eqref{eq:final_problem} is differentiable, a critical point can be found with the gradient method, whose iterations can then be written as
\begin{equation}
\theta_{t+1} = \operatorname*{argmin}_\theta \; \theta^\top \nabla U(\theta_t) + \frac{1}{2\alpha_t} \|\theta-\theta_t\|^2. 
\end{equation}
Replacing the Euclidean distance with a Bregman divergence $\mathbb{D}$ leads to the algorithm known as mirror descent
\begin{equation} \theta_{t+1} = \operatorname*{argmin}_{\theta} \; \theta^\top \nabla U(\theta_t) + \frac{1}{\alpha_t} \mathbb{D}(\theta,\theta_t). \label{eq:fgot_vmd} 
\end{equation}
For exponential-family distributions, the Kullback-Leibler (KL) divergence is a Bregman divergence \cite{raskutti2015information}, which is better indicated to optimize the parameters of a probability distribution \cite{Amari1998}:
\begin{equation}
\mathbb{D}(\theta,\theta_t) = \mathbb{D}_{\sf KL}(p_\theta \,||\, p_{\theta_t}) = \mathbb{E}_{P\sim p_\theta} \Big\{ \log\frac{p_\theta(P)}{p_{\theta_t}(P)} \Big\}.
\end{equation}

Among the possible choices for the distribution $p_\theta$, one of the most appropriate for our scenario is the diagonal Gaussian distribution $\mathcal{N}\big(\eta,\operatorname{diag}(\sigma^2)\big)$ with $\theta=(\eta,\sigma)$, because the KL divergence is given by
\begin{gather}
\mathbb{D}_{\sf KL}(p_{\eta,\sigma} \,||\, p_{\eta_t,\sigma_t}) = 
\qquad\qquad\qquad\qquad\qquad\qquad\qquad\nonumber\\
\qquad
\frac{1}{2} \sum_{d=1}^D \left( \frac{\sigma_d^2}{\sigma_{t,d}^2} + \frac{(\eta_{t,d}-\eta_d)^2}{\sigma_{t,d}^2} -1 + \log\frac{\sigma_{t,d}^2}{\sigma_{d}^2}\right).
\end{gather}
Then, 
the mirror-descent update boils down to
\begin{equation}
\begin{aligned}
\eta_{t+1,d} &= \eta_{t,d} - \alpha_t \sigma_{t,d}^2 \frac{\partial U(\theta_t)}{\partial \eta_d} \\[0.5em]
\sigma_{t+1,d} &= \sqrt{\sigma_{t,d}^2 + \big(\frac{\alpha_t\sigma_{t,d}^2}{2}\frac{\partial U(\theta_t)}{\partial \sigma_d}\big)^2} - \frac{\alpha_t\sigma_{t,d}^2}{2}\frac{\partial U(\theta_t)}{\partial \sigma_d}.
\end{aligned}
\end{equation}


The stochastic MGD algorithm for fGOT is summarised in Algorithm \ref{alg:alg_mgd}. Note that its computational complexity boils down to that of the matrix multiplications for each iteration of the algorithm. These can be computed very efficiently in the presence of sparse graphs, which is a common case in practice. We implemented Algorithm \ref{alg:alg_mgd} in PyTorch using the AMSGrad method \cite{j.2018on}. Our full implementation, using some functions from \cite{flamary2021pot}, is available at \url{https://github.com/Hermina/fGOT}.
\section{Experimental results}
\label{sec:fgot_exp}
We now illustrate the behaviour of fGOT in different experimental challenges. We consider three sets of experiments. The first one deals with random unstructured graphs, which we represent by random Erdos-Renyi graphs. We demonstrate the benefits of using high-pass filters to define a graph distance, and show the speed advancement compared to the GOT algorithm presented in \cite{NIPS2019_9539}. In the second set of experiments, we consider highly structured graphs, represented by stochastic block models. We tackle the problem of matching communities and show the benefits of low-pass filters. We also highlight the superior performance of our novel stochastic algorithm. Finally, we evaluate our method on a benchmark graph classification task, and compare to several state-of-the-art methods for graph alignment. The hyperparameters for all experiments were set empirically (for exact values, see the Appendix).

 


\subsection{Alignment of unstructured graphs}

\begin{figure}
	\begin{minipage}[b]{0.5\textwidth}
		\centering
		\includegraphics[width=0.99\textwidth]{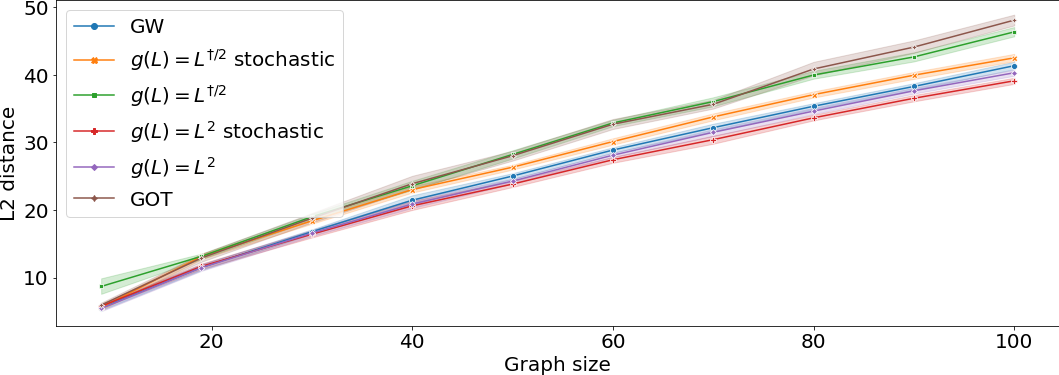}
	\end{minipage}  
	
	\caption{Performance comparison of different algorithms and filters on a task of random graph alignment. The performance is shown in terms of the Frobenius distance between aligned graph Laplacian matrices across different graph sizes.
		\label{fig:fgot_speed_l2}}
\end{figure}

\begin{figure}
	\begin{minipage}[b]{0.5\textwidth}
		\centering
		\includegraphics[width=0.99\textwidth]{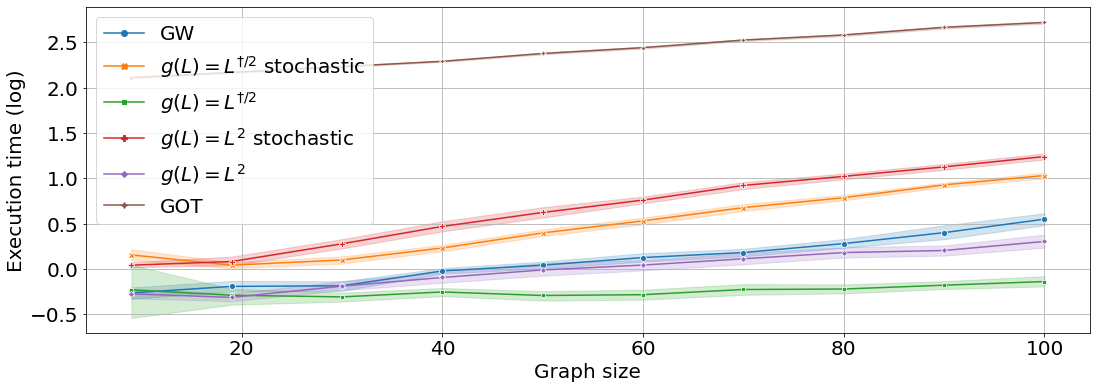}
	\end{minipage}    
	
	\caption{Speed comparison (natural log of execution time in \emph{seconds}) of different algorithms and filters on a task of random graph alignment across different graph sizes. 
		\label{fig:fgot_speed}}
\end{figure}
In our first experiment, we tackle the problem of alignment of unstructured random graphs across different graph sizes. We compare the alignment quality, as well as the execution time of different algorithms and filters. In particular, for each predefined graph size between 10 and 100, we perform 50 repetitions of aligning two random Erdos-Renyi graphs using the following algorithms: Gromov-Wasserstein (GW) \cite{2016-peyre-icml}, with graphs compared based on their shortest path matrices, as proposed in \cite{vayer2018optimal}; three algorithms solving for the original GOT cost: the original stochastic GOT algorithm, as proposed in \cite{NIPS2019_9539}, and our fGOT and stochastic fGOT algorithms, using the corresponding low pass filter $g(L) = \sqrt{L^{\dagger}}$; and two algorithms solving for a high-pass filter cost: fGOT and stochastic fGOT with $g(L) = L^2$. 

The task of unstructured graph alignment is better suited for distances defined by a high pass filter, as shown in Figure \ref{fig:fgot_speed_l2}. This is because high-pass filters prioritise local graph properties, which are crucial in the case of unstructured graphs due to the lack of global structure. 
 This reinforces the importance of the right filter choice for each problem, and emphasizes the benefits of the flexibility offered by fGOT.

\begin{figure}
	\begin{minipage}[b]{0.47\textwidth}
		\centering
		\includegraphics[width=0.99\textwidth]{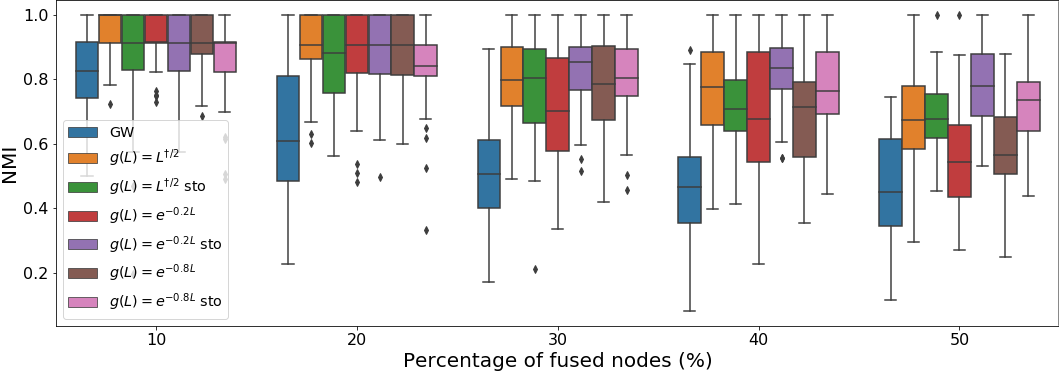}
	\end{minipage}    
	\vfill 
	\begin{minipage}[b]{0.47\textwidth}
		\centering
		\includegraphics[width=0.99\textwidth]{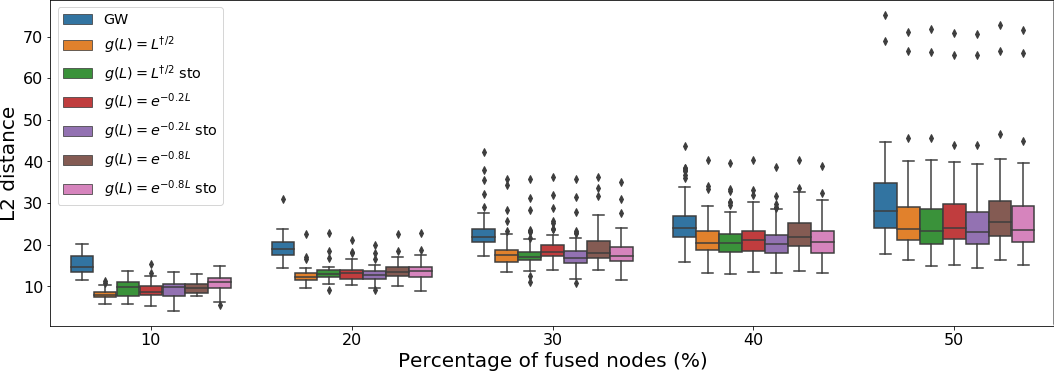}
	\end{minipage}  
	
	\caption{Alignment and community detection performance for distorted stochastic block model graphs as a function of the percentage of fused nodes (Experiment 1). The first plot shows the community detection performance in terms of NMI (closer to 1 is better), while the second one shows the Frobenius distance between aligned graph Laplacian matrices (closer to 0 is better).}
		\label{fig:fgot_community_distorted}
\end{figure}

Furthermore, we demonstrate the execution speed of each algorithm across different graph sizes. Compared to the original GOT formulation presented in \cite{NIPS2019_9539}, we can observe a significant speed improvement in both of our fGOT algorithms. In particular, using the corresponding filter $g(L) = \sqrt{L^{\dagger}} $, the MGD algorithm renders the computation of the original GOT distance comparable to the GW distance speed, circumventing the largest drawback of the GOT framework. 
The stochastic MGD algorithm also offers a great speed improvement, while successfully accounting for the nonconvexity of the alignment problem. This allows the algorithm to capitalise on both speed and accuracy (in terms of lower $l_2$ distance), offering better performance than its non-stochastic counterpart at a faster rate than GOT.

In summary, our two proposed algorithms offer a choice between maximising accuracy or speed, while both provide a very competitive trade-off between the two. Furthermore, our fGOT framework highlights the benefits of using filters in graph distances and offers high flexibility in choosing the appropriate filter for the problem at hand. For instance, unstructured graph alignment calls for high pass filters, while low-pass filters are more suitable for structured graphs, as we will see in the next section.

\subsection{Community detection in structured graphs}

We test the performance of fGOT on a graph alignment and community detection task in structured graphs. We consider two experimental settings: the comparison of a graph with a noisy version of the same graph, and its comparison with a random structured graph. 
We show the results obtained by GW (as described in the previous section) and by the MGD and the stochastic MGD version of fGOT for three different graph filter models. Namely $g(L) = \sqrt{L^{\dagger}}$, which corresponds to the approximated version of the original GOT distance for smooth signals \cite{NIPS2019_9539}, and two heat kernel distances: $g(L) = e^{-0.2 L}$ and $g(L) = e^{-0.8 L}$.

\begin{figure}
	\begin{center}
		\begin{minipage}[b]{0.47\textwidth}
			\centering
			\includegraphics[width=0.99\textwidth]{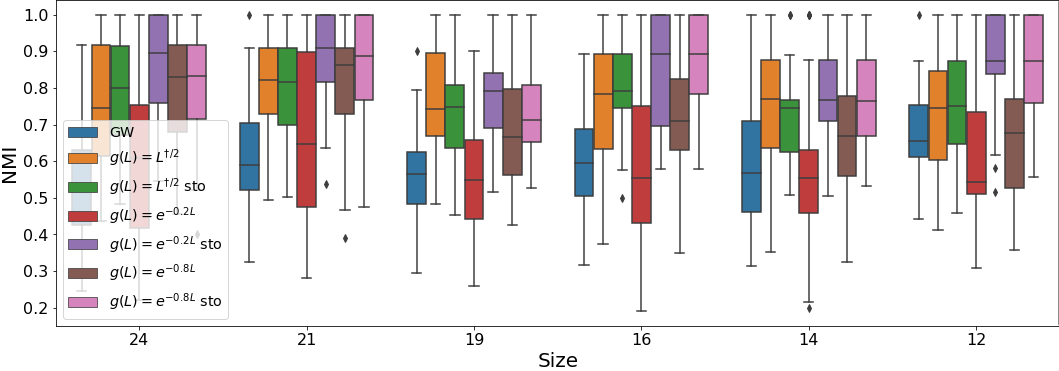}
		\end{minipage}    
		\vfill 
		\begin{minipage}[b]{0.47\textwidth}
			\centering
			\includegraphics[width=0.99\textwidth]{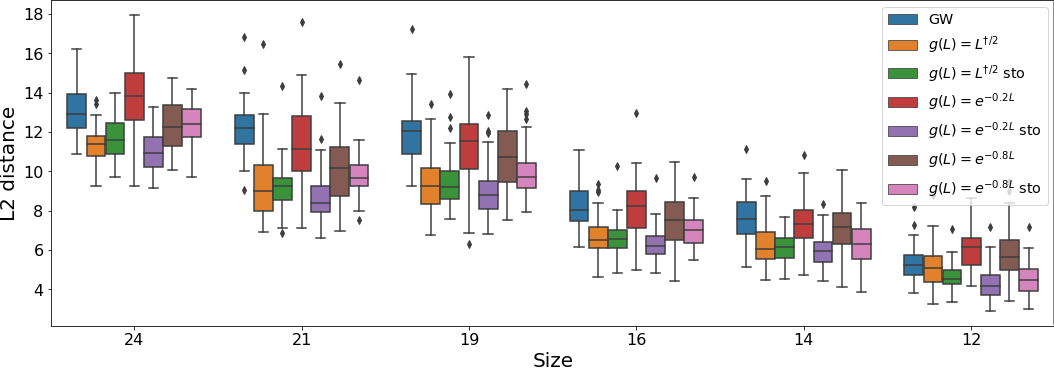}
		\end{minipage}  
	\end{center}
	\caption{Alignment and community detection performance for random instances of stochastic block model graphs as a function of the graph size (Experiment 2). The first plot shows the community detection performance in terms of NMI (closer to 1 is better), while the second one shows the Frobenius distance between aligned graph Laplacian matrices (closer to 0 is better).}
	\label{fig:fgot_community_size}
\end{figure}

 \begin{table*}[h!]
 \caption{1-NN classification accuracy}
  \label{tab:classification_copt}
 \centering
 \footnotesize	
 \begin{tabular}{|l|l|l|l|l|l|}
 \hline
 & \bf BZR                    & \bf MUTAG                 & \bf ENZYMES               & \bf PROTEIN               & \bf AIDS                  \\ \hline
 $g_1(L) = \sqrt{L\dagger}$ & $ 77.7  \pm  7.17  $   & $ 85.56  \pm  5.98  $ & $ 27.89  \pm  9.01  $ & $ 64.56  \pm  9.56  $ & $ 99.0  \pm  1.95  $  \\ \hline
 $g_2(L) = L^2 $  & $ \bf 82.07  \pm  6.34  $  & $ 84.56  \pm  6.69  $ & $ 26.0  \pm  8.71  $  & $ 61.0  \pm  9.32  $  & $ 96.89  \pm  3.44  $ \\ \hline
 $g_3(L) = e^{-0.8 L}$  & $ 79.43  \pm  7.17  $  & $ \bf 86.78  \pm  5.87  $ & $ 25.56  \pm  7.06  $ & $ 62.44  \pm  9.92  $ & $ 98.78  \pm  2.02  $ \\ \hline
 $g_4(L) = \sqrt{L\dagger} + L^2 $  & $ 78.05  \pm  11.36  $ & $ 84.44  \pm  6.8  $  & $ 24.33  \pm  9.89  $ & $ \bf 66.22  \pm  9.69  $ & $ \bf 99.44  \pm  1.24  $ \\ \hline
 $g_5(L) = \sqrt{L\dagger} + e^{-0.8 L}$  & $ 78.39  \pm  6.53  $  & $ 86.44  \pm  5.37  $ & $ \bf 29.0  \pm  8.74  $  & $ 64.67  \pm  9.13  $ & $ 99.22  \pm  1.41  $ \\ \hline
 $g_6(L) = L^2 + e^{-0.8 L}$  & $ 77.01  \pm  10.95  $ & $ 83.22  \pm  7.58  $ & $ 26.67  \pm  7.2  $  & $ 63.56  \pm  8.47  $ & $ \bf 99.44  \pm  1.51  $ \\ \hline
 GW  & $ 79.22  \pm  7.68  $  & $ 84.11  \pm  6.98  $ & $ 24.67  \pm  5.94  $ & $ 61.22  \pm  9.33  $ & $ 95.0  \pm  4.28  $  \\ \hline
 GA &  $72.11 \pm9.69 $       & $70.33 \pm8.09 $      & $17.44 \pm6.17 $      & $47.89 \pm11 $        & $82.56 \pm8.29 $      \\ \hline
 IPFP & $69 \pm8.54 $          & $77.89 \pm6.92 $      & $21.22 \pm6.34 $      & $49.67 \pm7.5 $       & $84.22 \pm7.68 $      \\ \hline
 RRWM   & $66.11 \pm12.66 $      & $64.44 \pm10.55 $     & $20.44 \pm8.292 $     & $47.22 \pm11.58 $     & $81.33 \pm8.824 $     \\ \hline
 \end{tabular}
 \end{table*}
 
To evaluate the alignment recovery, we resort to two measures: Normalised mutual information (NMI) of community alignment, and the difference between the aligned graphs in terms of the $\ell_2$ norm. We estimate the community alignment of these structured graphs by performing spectral clustering in both (aligned) graphs, separating their nodes into 4 communities. The two measures can be considered as complementary, with NMI capturing the recovery of global structure, and $\ell_2$ considering local differences independently. 

In Experiment 1, we generate a stochastic block model graph $\mathcal{G}_2$ with 24 nodes and 4 communities. The graph $\mathcal{G}_1$ is constructed as a noisy version of $\mathcal{G}_2$ by randomly collapsing edges until a target percentage of nodes is fused. Graph $\mathcal{G}_1$ is then further randomly permuted, in order to change the order of its nodes. Each experiment has been repeated 50 times with a different $\mathcal{G}_2$. The results are given in Figure \ref{fig:fgot_community_distorted}. 

In Experiment 2, both graphs are generated randomly as instances of stochastic block models with 4 communities. For each $\mathcal{G}_2$ with size 24, we generate six graphs $\mathcal{G}_1$ with a different number of edges and vertices (between 12 and 24). While we keep the number of communities equal for a meaningful community detection task, there is no other direct connection between $\mathcal{G}_1$ and $\mathcal{G}_2$. Each experiment was repeated 50 times with a different $\mathcal{G}_2$. The results are in Figure \ref{fig:fgot_community_size}.

Both experiments demonstrate the superior performance of our stochastic algorithm compared to the non-stochastic MGD algorithm, especially in the case of heat kernel models. The improvements are especially visible in the community detection tasks, as the settings become more challenging. 
Furthermore, the stochastic version of the heat kernel distances outperforms both the approximated GOT distance, as well as the GW in both experiments. This shows that, even when observing only smooth filters, the flexibility offered by considering different filter models can affect our results substantially and bring considerable benefits.
Finally, the results in terms of the $l_2$ distance are consistent with the NMI, suggesting that the stochastic algorithm improves the alignment significantly both in terms of global structure recovery and more local alignment properties.

\subsection{Graph classification}

Finally, we test the effect of filter graph distances in a graph classification task. We explore five different benchmark datasets: BZR dataset (average number of nodes $\bar{N}$ = 35.75)  \cite{sutherland2003spline}, MUTAG ($\bar{N}$= 17.93) \cite{debnath1991structure}, ENZYMES ($\bar{N}$ = 32.63), PROTEIN ($\bar{N}$ = 39.06) \cite{borgwardt2005shortest} and AIDS ($\bar{N}$= 15.69). From each dataset, we sample 100 graphs, use fGOT to align them and to compute normalised pairwise distances, and then perform a simple non-parametric 1-NN classification. We repeat this experiment 30 times and report the result in terms of the mean and standard deviation of the classification accuracy. We compare several variants of fGOT, emphasising the flexibility of the method and the benefits this flexibility can bring when comparing datasets of different nature. The filters in comparison are as follows: $g_1(L) = \sqrt{L\dagger}$, $g_2(L) = L^2$, $g_3(L) = e^{-0.8 L}$, $g_4(L) = \sqrt{L\dagger} + L^2$, $g_5(L) = \sqrt{L\dagger} + e^{-0.8 L}$, $g_6(L) = L^2 + e^{-0.8 L}$. These variants of fGOT are further compared to several state-of-the-art graph alignment algorithms: GW \cite{2016-peyre-icml, vayer2018optimal}, GA \cite{gold1996graduated}, IPFP \cite{leordeanu2009integer} and RRWM \cite{cho2010reweighted}. As shown in Table \ref{tab:classification_copt}, a high-pass filter (eg. $g_2(L)$) will have a higher classification accuracy on a dataset with local changes, such as BZR. Conversely, a low-pass filter (eg. $g_1(L), g_3(L)$) will perform better in a dataset with strong global differences, such as AIDS. This reinforces the value of the flexibility that fGOT offers, and allows us to choose the right distance criteria for our data. 
Even in the more challenging task of 6-class classification (ENZYMES), the fGOT distances seem to be more relevant, even though we note that 1-NN classification is probably too simplistic to solve such difficult tasks. Finally, composite filters ($g_4(L), g_5(L), g_6(L)$), covering both low and high frequencies, perform competitively on all observed datasets, suggesting their efficiency in retaining different captured effects. This makes them general enough to use on data for which we might not have a clear filter preference, or when we are unaware of the nature and properties of our graphs. 

\section{Conclusion}
\label{sec:fgot_conclusion}
We proposed fGOT, a flexible method for graph comparison based on optimal transport, which uses filter models to encode specific graph properties. We exploit the representation of graphs through the distribution of filtered signals, which permits an efficient comparison through the Wasserstein distance of these distributions. In order to provide a more scalable algorithm, we formulate an approximation to the generic filter optimal transport cost, circumventing the computation of its most expensive parts. We propose an efficient stochastic algorithm based on Bayesian exploration, which adapts mirror gradient descent to this challenging non-convex problem. We show the performance of our method in the context of graph alignment, community detection and classification. Experimental results show that both our proposed algorithms offer a superior performance in terms of speed, when compared to the original GOT algorithm. Furthermore, our novel stochastic algorithm offers superior performance to its non-stochastic counterpart in terms of accuracy and alignment quality, reaching a very good trade-off between speed and accuracy. Finally, experiments on both unstructured and structured graphs show that the filter graph distance brings valuable flexibility, with problem-adjusted filters leading to better performance than classical metrics in different simple tasks. 

\bibliography{aaai22}

\newpage

\appendix
\section{Appendix}

\subsection{Proofs of Lemmas}
\begin{proof}[Proof of Lemma \ref{lemma_1}]
	We write the eigenvalue decomposition of $L$ as $L = U \Lambda U^T$. Notice that the eigenvalue decomposition of $PLP^T$ is then $(PU) \Lambda (PU)^T$. Namely, with $u_i$ an eigenvector of $L$ corresponding to the eigenvalue $\lambda_i$, we have:
	\small{\begin{equation}
		PLP^T Pu_i = PLu_i = \lambda_i Pu_i
	\end{equation}}
	Therefore,
	\small{\begin{align}
		g(PLP^T) & = g((PU) \Lambda (PU)^T) = (PU) \hat{G} (PU)^T \nonumber \\ & = P U \hat{G} U^T P^T = Pg(L)P^T
	\end{align}}
\end{proof}
\begin{figure*}[h!]
	\centering
	\includegraphics[width=0.9\textwidth]{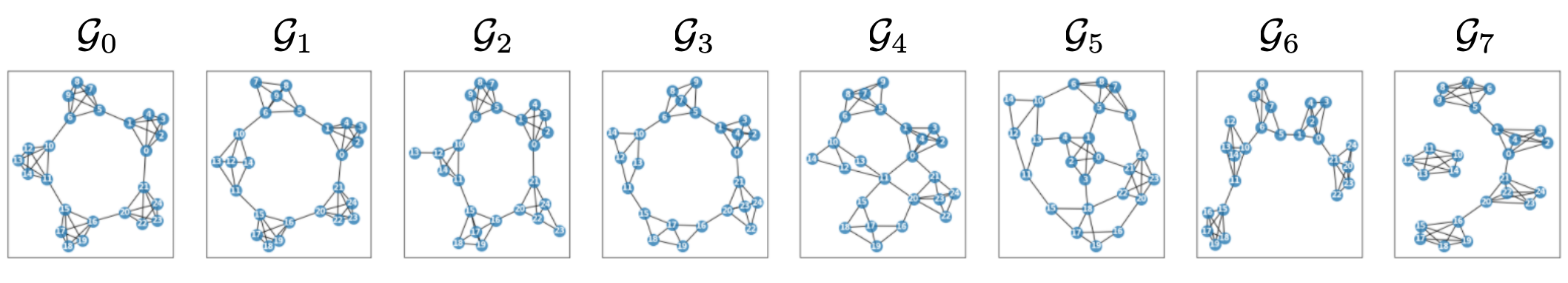}
	\caption{Enumerated list of random graphs that are used as illustrative examples for graph comparison. Graphs are generated by randomly perturbing the original graph $\mathcal{G}_0$ with adding and removing random edges.}
	\label{fig:fgot_examples_list}
\end{figure*}

\begin{proof}[Proof of Lemma \ref{lemma_2}]
{\small{
	\begin{align}
		\underbrace{\mathcal{W}_2^2\big(\nu^{\mathcal{G}_1, g}, \nu^{\mathcal{G}_2, g}_{P^T}\big)}_\mathcal{J} 
		&= {\rm Tr}\left(g^2(L_1) \right) + {\rm Tr}\left(P g^2(L_2) P^T\right) \nonumber \\ & \hspace{3mm} - 2 \, {\rm Tr}\left(\sqrt{g(L_1)P g^2(L_2) P^T g(L_1)}\right)\\
		&= {\rm Tr}\left(g^2(L_1) \right) + {\rm Tr}\left(g^2(L_2) P^T P\right) \nonumber \\ & \hspace{3mm} - 2 \sum_i { \lambda_i}\left(\sqrt{g(L_1)P g^2(L_2) P^T g(L_1)}\right)\label{eq:proof_trace_eig} \\
		&= {\rm Tr}\left(g^2(L_1) \right) + {\rm Tr}\left(g^2(L_2) P^T P\right) \nonumber \\ & \hspace{3mm} - 2 \sum_i \sqrt{ {\lambda_i} \left( g(L_1)P g^2(L_2) P^T g(L_1) \right) }\\
		&= {\rm Tr}\left(g^2(L_1) \right) + {\rm Tr}\left(g^2(L_2) \right) \nonumber \\ & \hspace{3mm} - 2 \sum_i \sqrt{ {\lambda_i} \left( g(L_1)P g(L_2) P^T P g(L_2) P^T g(L_1) \right) }\label{eq:proof_lambda}
	\end{align}
}}
	with $\lambda_i(A)$ denoting the eigenvalues of a matrix $A$. Here, (\ref{eq:proof_trace_eig}) follows because for a symmetric positive-semidefinite matrix $Tr(A) = \sum_i \lambda_i(A)$ and (\ref{eq:proof_lambda}) because $PP^T = P^TP = I_N$. Now with $\sigma_i(A)$ denoting the singular values of a matrix $A$, the equality becomes:
{\small{
	\begin{align}
		\mathcal{J} &= {\rm Tr}\left(g^2(L_1) \right) + {\rm Tr}\left(g^2(L_2) \right) \nonumber \\ & \hspace{3mm} - 2 \sum_i \sqrt{ {\lambda_i} \left(\left( g(L_1)P g(L_2) P^T \right) \left( g(L_1)P g(L_2) P^T \right)^T \right) }\\
		&= {\rm Tr}\left(g^2(L_1) \right) + {\rm Tr}\left(g^2(L_2) \right) \nonumber \\ & \hspace{3mm} - \sum_i \left({\sigma_i} \left( g(L_1)P g(L_2) P^T \right) + {\sigma_i} \left( \left( g(L_1)P g(L_2) P^T \right)^T \right) \right), \label{eq:proof_sigma}
	\end{align}
}}
	because $\lambda_i(AA^T) = \sigma_i^2(A) = \sigma_i^2(A^T)$. Finally, (\ref{eq:sigma_ineq}) follows from $\sum _i \sigma_i(A + B) \leq \sum_i (\sigma_i(A) + \sigma_i(B))$, and (\ref{eq:sigma_lambda}) because for a symmetric positive-definite matrix $C$, $\sigma_i(C) = \lambda_i(C)$:
	\small{\begin{align}
		\mathcal{J} &\leq {\rm Tr}\left(g^2(L_1) \right) + {\rm Tr}\left(g^2(L_2) \right) \nonumber \\ & \hspace{3mm} - \sum_i \left({\sigma_i} \left( g(L_1)P g(L_2) P^T  + \left( g(L_1)P g(L_2) P^T \right)^T \right)\right)\label{eq:sigma_ineq}\\
		& = {\rm Tr}\left(g^2(L_1) \right) + {\rm Tr}\left(g^2(L_2) \right) \nonumber \\ & \hspace{3mm}  - \sum_i {\lambda_i} \left( g(L_1)P g(L_2) P^T  + \left( g(L_1)P g(L_2) P^T \right)^T \right)\label{eq:sigma_lambda}\\
		&= {\rm Tr}\left(g^2(L_1) + g^2(L_2) \right) \nonumber \\ & \hspace{3mm}  - {\rm Tr} \left( g(L_1)P g(L_2) P^T  + \left( g(L_1)P g(L_2) P^T \right)^T\right)\\
		&= {\rm Tr}\left(g^2(L_1) + g^2(L_2) \right) - 2 \, {\rm Tr} \left( g(L_1)P g(L_2) P^T \right)\\
		&= {\rm Tr}\left(g^2(L_1) + g^2(L_2) \right)- 2\left\langle g(L_1)P g(L_2),  P \right\rangle\\
		&= \widetilde{\mathcal{W}}_2^2\big(\nu^{\mathcal{G}_1, g}, \nu^{\mathcal{G}_2, g}_{P}\big)
	\end{align}}
\end{proof}

\begin{figure*}[h!]
	\centering
	\includegraphics[width=\textwidth]{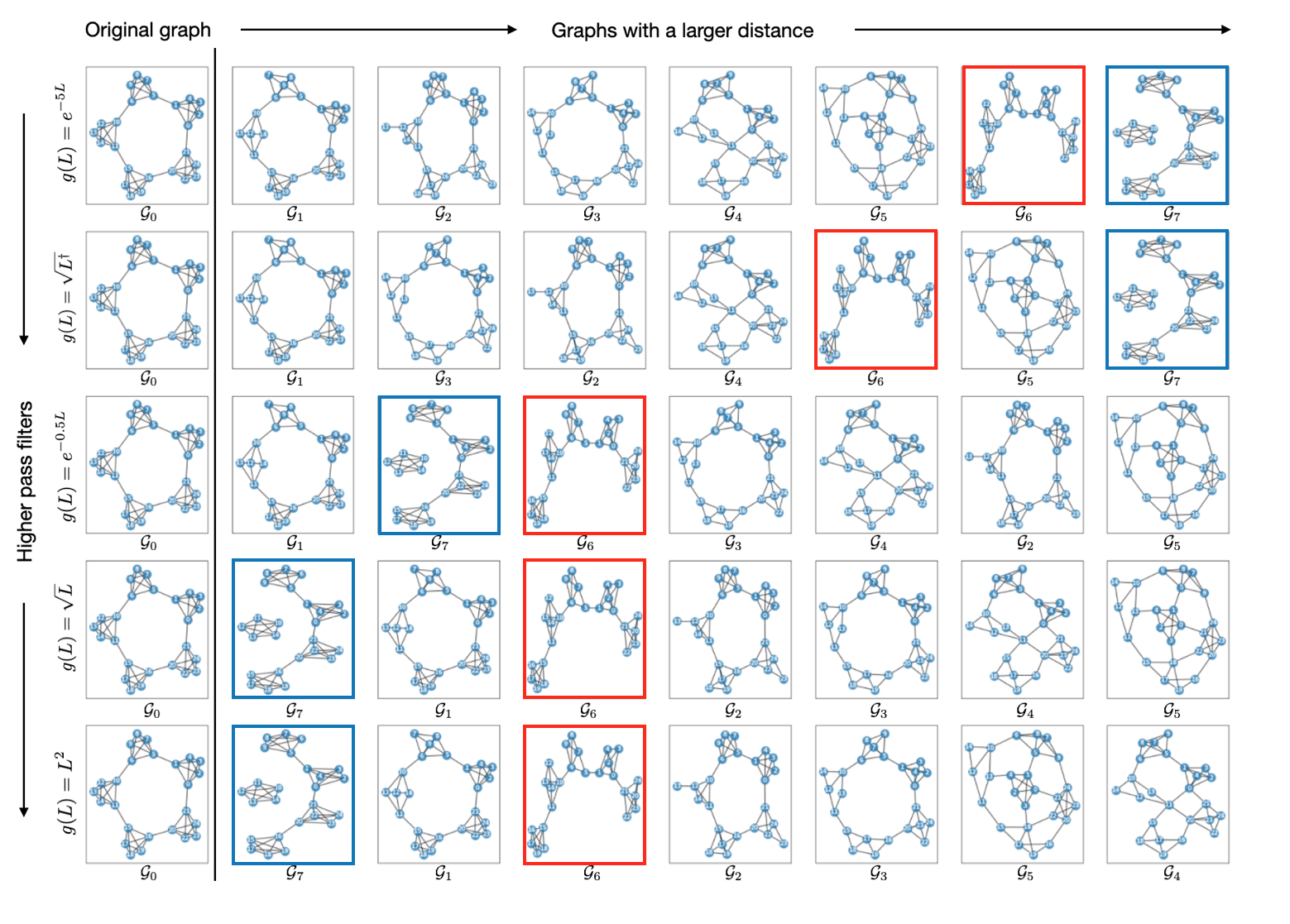}
	\caption{Random graphs of Figure \ref{fig:fgot_examples_list} sorted based on increasing distance to $\mathcal{G}_0$, for different filter graph distances. Each row is a different filter distance, starting with very smooth filters and going towards more high pass filter distances. Namely, the filters are given by $g_1(L) = e^{-5L}, g_2(L) = \sqrt{L^{\dagger}}, g_3(L) = e^{-0.5L}, g_4(L) = \sqrt{L}, g_5(L) = L^2$.}
	\label{fig:fgot_examples}
\end{figure*}

\begin{proof}[Proof of Lemma \ref{lemma_3}]
Note that both $\mathcal{W}_2^2$ and $\widetilde{\mathcal{W}}_2^2$ are non-negative by definition. As $\mathcal{G}_1$ and $\mathcal{G}_2$ are isomorphic, there exists a $P\in \mathcal{C}_{\rm perm}$ such that $L_1 = P L_2 P^T$. For that $P$, we have:
\small{\begin{align}
    \mathcal{W}_2^2\big(\nu^{\mathcal{G}_1, g}, \nu^{\mathcal{G}_2, g}_{P^T}\big)
    &= {\rm Tr}\left(g^2(L_1) \right) + {\rm Tr}\left(P g^2(L_2) P^T\right) \nonumber \\ & \hspace{3mm} - 2 \, {\rm Tr}\left(\sqrt{g(L_1)P g^2(L_2) P^T g(L_1)}\right)\\
    &= {\rm Tr}\left(g^2(L_1) \right) + {\rm Tr}\left(g^2(L_1)\right) \nonumber \\ & \hspace{3mm} - 2 \, {\rm Tr}\left(\sqrt{g(L_1)g^2(L_1) g(L_1)}\right)\\
    &= 2 \, {\rm Tr}\left(g^2(L_1) \right) - 2 \, {\rm Tr}\left(g^2(L_1)\right)\\
    &= 0.
\end{align}
\begin{align}
    \widetilde{\mathcal{W}}_2^2\big(\nu^{\mathcal{G}_1, g}, \nu^{\mathcal{G}_2, g}_{P}\big) 
    &= {\rm Tr}\left(g^2(L_1) + g^2(L_2) \right) \nonumber \\ & \hspace{3mm} - 2\left\langle g(L_1)P g(L_2),  P \right\rangle\\
    &= {\rm Tr}\left(g^2(L_1)\right) + {\rm Tr}\left(P g^2(L_2) P^T\right) \nonumber \\ & \hspace{3mm} - 2 \, {\rm Tr} \left( g(L_1)P g(L_2) P^T \right)\\
    &= 2 \, {\rm Tr}\left(g^2(L_1)\right) - 2 \, {\rm Tr}\left(g^2(L_1)\right) \\
    &= 0.
\end{align}}
\end{proof}

\subsection{Hyperparameters used in experiments}
Let $n$ be the size of $\mathcal{G}_1$ and $m$ be the size of $\mathcal{G}_2$. Further, let $max1$ be the maximum element of $g(L_1)$ and $max2$ be the maximum element of $g(L_2)$. In the alignment of unstructured graphs, the hyperparameters were set as follows: for MGD $g(L) = \sqrt{L^{\dagger}}$ to $\epsilon = \frac{6\times 10^{-3}max1 max2}{\sqrt{nm}}$; for MGD $g(L) = L^2$ to $\epsilon = \frac{3\times 10^{-3}max1 max2}{\sqrt{nm}}$; for all versions of stochastic MGD to $\tau = 1$, samples = 5, $\alpha = \frac{50nm}{max1 max2}$.

In the community detection in structured graphs, the hyperparameters were set as follows: for MGD with both perturbed and random graphs $g(L) = \sqrt{L^{\dagger}}$ to $\epsilon = \frac{8\times 10^{-3}max1 max2}{\sqrt{nm}}$; for $g(L) = e^{-0.2 L}$ and $g(L) = e^{-0.8 L}$ to $\epsilon = \frac{2\times 10^{-2}max1 max2}{\sqrt{nm}}$; for all versions of the stochastic MGD we set $\tau = 1$, samples = 5, $\alpha = \frac{nm}{max1 max2}$.


In graph classification, the hyperparameters were set as follows: for all filters, a simple line search for parameter $\epsilon$ was performed prior to conducting the experiments, and the best parameters were automatically selected and used in subsequent experiments. For every graph pair, the parameter $\epsilon$ was normalised with $\frac{max1max2}{\sqrt{nm}}$. 

When using our algorithms, for the non-stochastic fGOT version we recommend setting $\epsilon = c_1 \times \frac{max1max2}{\sqrt{nm}}$.
For the stochastic fGOT version, we recommend fixing $\tau = 1$, samples = 5, and setting $\alpha = c_2 \times \frac{nm}{max1max2}$.
The $c_1$ and $c_2$ values can be fixed with a line search or taken as empirical values reported in this section.

\subsection{An illustration of filter graph distances}
To illustrate the effect of filters on the definition of a graph distance, we compare the set of graphs presented in Figure \ref{fig:fgot_examples_list} based on different filter distances, and sort them based on their distance to $\mathcal{G}_0$. Figure \ref{fig:fgot_examples} shows the flexibility of fGOT in prioritising different phenomena in the definition of our distance. There are several differences in the ordering of graphs with different filter distances. For instance, $\mathcal{G}_6$ exchanges places with $\mathcal{G}_2$ and $\mathcal{G}_3$ as the filter becomes high pass, from rows 1 and 2 to rows 4 and 5. The reason for this is that smooth filters capture the rupture in the global ring structure of $\mathcal{G}_6$, while the higher pass filters focus on local changes, which are more present in $\mathcal{G}_2$ and $\mathcal{G}_3$. The same example shows the strong impact of temperature on the behaviour of the heat kernel filter.
Namely, $g_1(L) = e^{-5L}$ has a very large reach, which makes it the smoothest filter we observe, while the very limited reach of $g_3(L) = e^{-0.5L}$ makes it more focused on local changes, and positions it between the low and high pass filters in this example. 

Finally, we note that a filter graph distance can be especially useful for the systematic comparison of graphs with disconnected components. Namely, traditional distances comparing Laplacian or Adjacency matrices directly will not take any connectedness information into account. More meaningful distances like GOT or GW are not designed for disconnected graphs, and need to use heuristic solutions in order to compare those. At the same time, a filter graph distance can easily control the importance of connectedness by adjusting its spectral properties. An example of this can be seen in row 3 of Figure \ref{fig:fgot_examples} with $g_3(L) = e^{-0.5L}$ and graph $\mathcal{G}_7$, where the connectedness information is taken into account without being given too much importance. 
This can be understood intuitively from the spectral perspective: a disconnected graph will have the second eigenvalue $\lambda_2 = 0$. Therefore, $\hat{g}_3(\lambda_2)$ for a small $\lambda_2$ in a connected graph will be significantly different from $\hat{g}_3(0)$, resulting in a reasonably large contribution towards the overall graph distance. At the same time, this contribution is clearly bounded and well defined, leaving enough room for other differences between graphs to be taken into account.

\end{document}